\pdfoutput=1
\def\year{2021}\relax
\documentclass[letterpaper]{article} % DO NOT CHANGE THIS
\usepackage{aaai21}  % DO NOT CHANGE THIS
\usepackage{times}  % DO NOT CHANGE THIS
\usepackage{helvet} % DO NOT CHANGE THIS
\usepackage{courier}  % DO NOT CHANGE THIS
\usepackage[hyphens]{url}  % DO NOT CHANGE THIS
\usepackage{graphicx} % DO NOT CHANGE THIS
\usepackage{natbib}  % DO NOT CHANGE THIS AND DO NOT ADD ANY OPTIONS TO IT
\usepackage{caption} % DO NOT CHANGE THIS AND DO NOT ADD ANY OPTIONS TO IT
\usepackage[switch]{lineno} 
\usepackage{hyperref} 
\usepackage{footnote}
\usepackage{multirow}
\usepackage{multicol}
\usepackage{graphicx}
\usepackage{amsmath,amssymb} % define this before the line numbering.
\usepackage{color}
\usepackage{dsfont}
\usepackage{subcaption}

\renewcommand{\arraystretch}{1.3}
\title{Boosting Image-based Mutual Gaze Detection using Pseudo 3D Gaze}
\author{
    Bardia Doosti\thanks{This work was done when Bardia Doosti was interning at Google.}$^{1}$, Ching-Hui Chen\thanks{Ching-Hui Chen is currently affiliated with Waymo.}$^{2}$, Raviteja Vemulapalli$^{2}$, \\ Xuhui Jia$^{2}$, Yukun Zhu$^{2}$, Bradley Green$^{2}$ 
    \\
}
\affiliations{
Luddy School of Informatics, Computing, and Engineering, Indiana University Bloomington$^{1}$ \\    
bdoosti@indiana.edu \\
    Google Research$^{2}$ \\
    \{chuichen,
ravitejavemu,xhjia,yukun,brg\}@google.com \\

}
\begin{document}
%\linenumbers 
\maketitle

\begin{abstract}
Mutual gaze detection, i.e., predicting whether or not two people are looking at each other, plays an important role in understanding human interactions. In this work, we focus on the task of image-based mutual gaze detection, and propose a simple and effective approach to boost the performance by using an auxiliary 3D gaze estimation task during the training phase. We achieve the performance boost without additional labeling cost by training the 3D gaze estimation branch using pseudo 3D gaze labels deduced from mutual gaze labels. By sharing the head image encoder between the 3D gaze estimation and the mutual gaze detection branches, we achieve better head features than learned by training the mutual gaze detection branch alone. Experimental results on three image datasets show that the proposed approach improves the detection performance significantly without additional annotations. This work also introduces a new image dataset that consists of 33.1K pairs of humans annotated with mutual gaze labels in 29.2K images.
\end{abstract}

\section{Introduction}
With recent advances in the field of computer vision, social relation discovery has received significant attention from the research community~\cite{ding2010learning,ding2011inferring,Liu2019,sapru2015automatic,sun2017domain,wang2010seeing,yu2009monitoring}. Detecting interactions such as hugging, kissing, mutual gaze, etc., can aid machines in understanding people's social behavior since such cues often convey significant information about a person's social intentions. Among all verbal and non-verbal cues, gaze is known to be one of the most informative cues about social interactions. Existing works have shown that gaze can be used as a reliable proxy for identifying the social relationships between people in a scene~\cite{fan2018inferring,fan2019understanding,Marin19a}. 
In particular, mutual eye contact between two people in an image often indicates social connectivity, friendliness and intimacy~\cite{abele1986functions}.

In this work, we focus on the task of image-based mutual gaze detection and propose a deep learning-based approach that directly predicts mutual gaze score for a pair of head bounding boxes in an image. First, each head image patch is converted into a compact feature representation using a Convolutional Neural Network (CNN). Then, the spatial relationship between the head bounding boxes is encoded using 2D and geometry-inspired 3D spatial features. Specifically, our 3D spatial feature is the direction of the relative 3D position vector between the two head bounding box centers. Finally, the head spatial and image encoded features are concatenated and processed by a fully-connected neural network that outputs mutual gaze score. This is our primary mutual gaze detection branch.

We also use an auxiliary 3D gaze estimation branch during training. This branch takes the head image embeddings as inputs and estimates the 3D gaze vectors of the corresponding heads. The mutual gaze detection and 3D gaze estimation branches are trained jointly with a shared head image encoder in an end-to-end fashion to take advantage of multi-task learning. The auxiliary gaze estimation branch helps in training a better head image encoder which in turn leads to improved performance of the mutual gaze detection branch. Though the proposed model has a 3D gaze estimation branch, we do not use any 3D gaze ground truth for training. Instead, we use the direction of the relative 3D head position vector as the pseudo 3D gaze label for positive mutual gaze samples. We use these pseudo labels for training and improve mutual gaze detection performance without additional labeling cost. Our experimental results on three mutual gaze image datasets show that the proposed approach outperforms the state of the art.\\

\noindent\textbf{Major contributions:}
\begin{itemize}
\itemsep 4pt
    \item We propose to use an auxiliary 3D gaze estimation task along with the primary mutual gaze detection task.
    %to boost the performance of mutual gaze detection.
    We show that the mutual gaze detection performance can be improved without additional labeling by training the auxiliary 3D gaze estimation branch using pseudo 3D gaze labels generated for positive mutual gaze samples.
    \item We propose to use the direction of the relative 3D head position vector both as pseudo 3D gaze label and as an input feature for mutual gaze detection. We show how this 3D vector can be computed approximately without having ground truth depth.
    \item We introduce a new in-the-wild image dataset that consists of 33.1K pairs of humans annotated with mutual gaze in 29.2K images. This dataset can be downloaded from \textcolor{magenta}{ \href{https://research.google/tools/datasets/google-open-images-mutual-gaze-dataset/}{https://research.google/tools/datasets/google-open-images-mutual-gaze-dataset/}}. 
    \item We show that the proposed approach outperforms the state-of-the-art mutual gaze detection approach of~\cite{Marin19a}~\footnote{Since this work focuses on image-based detection, we convert the original video-based LAEO-Net into an image-based LAEO-Net by replacing 3D convolutions with 2D convolutions.} on three image datasets.
\end{itemize}

\section{Related Work}
This work intersects with three topics in the field of  computer vision: visual focus of attention prediction, 3D gaze estimation, and mutual gaze detection. 
\subsection{Visual focus of attention}
Visual Focus of Attention (VFA) prediction aims at identifying where people in an image are looking at within the image space~\cite{ba2008recognizing}.
\cite{nips15_recasens} proposed a two-stream CNN to find the position where people in the image are looking at. The first stream predicts a saliency heatmap from the image, and the second stream predicts a gaze heatmap in the image space from the head image. The VFA heatmap is then computed by the element-wise product of saliency heatmap and gaze heatmap. \cite{Chong2018} leveraged the looking inside/outside the image annotation as additional supervision to improve the performance of VFA prediction. Since eyes are not always visible, \cite{Mass2019} exploited the correlation between eye gaze and head movements for tracking the gaze and VFA.
\cite{fan2018inferring} defined \textit{shared attention} as a phenomenon where two or more individuals simultaneously look at a common target in social scenes. Identifying shared attention of people in videos provides cues for understanding their social activity and identifying their social intent. They proposed a network architecture for spatio-temporal modeling of shared attention by taking the gaze heatmap and region proposal heatmap as inputs. The gaze heatmap provides the spatial information where a group of people are looking at, and the region proposal heatmap provides the context of objects/targets in the scene.

\subsection{3D gaze estimation}
As 3D gaze direction of a person is an important cue for understanding human attention, 3D gaze estimation task has received significant attention from the computer vision community~\cite{Brau2018,huang2017screenglint,krafka2016eye,zhang2015appearance,zhang2017s,zhu2017monocular}. \cite{fischer2018rt} estimated the 3D gaze direction from image patches of face and eyes. The 3D gaze direction data was collected using a motion capture system (eight cameras), RGB-D camera, and eye-tracking glasses. \cite{Zhang2018mpii} collected a 3D gaze estimation dataset from 15 users during their laptop usage over several months. Recently, \cite{Kellnhofer2019} introduced a large-scale Gaze360 dataset which was collected using a $360^{\circ}$ camera that simultaneously captured both the subject and a mobile gaze fixation target.

Creating a large-scale dataset with 3D gaze ground truth requires significant amount of manual effort, and 3D gaze estimation models trained on a small set of individuals may not generalize well. Though we use 3D gaze estimation as an auxiliary task in our approach, we do not use any 3D gaze ground truth for training. Instead, we generate pseudo 3D gaze labels for people looking at each other, and use these pseudo labels for training. This way, we boost the performance of mutual gaze detection without any additional labeling cost.

\subsection{Mutual gaze detection}
Mutual gaze is one of the most common types of social interaction.
\cite{Marin11,marin2014detecting} proposed to detect mutual gaze by modeling the head poses and spatial relationship in the scene.
Recently, \cite{palmero2018automatic} built a mutual gaze detection model on top of 3D gaze vectors estimated using a CNN and the locations of faces in the image.
Instead of explicitly estimating the 3D gaze and inferring mutual gaze from it, \cite{Marin19a} proposed LAEO-Net for directly detecting mutual gaze in videos. LAEO-Net consists of two CNNs for encoding head images and head location map, respectively. The encoder utilizes 3D convolutions for temporal modeling, and the encoded features are concatenated and provided as input to a binary classifier. To understand the human gaze communication among a group of people, \cite{fan2019understanding} proposed a spatio-temporal graph modeling approach for predicting various types of human gaze communication (e.g., single/mutual gaze, shared gaze, etc.) in social scene videos. In contrast to these works which focus on video-based mutual gaze detection, this work focuses on single image-based mutual gaze detection.

\section{Proposed Approach}
\begin{figure*}[t]
\begin{center}
   \includegraphics[trim={0cm 0.10cm 0cm 0cm}, width =0.8\textwidth]{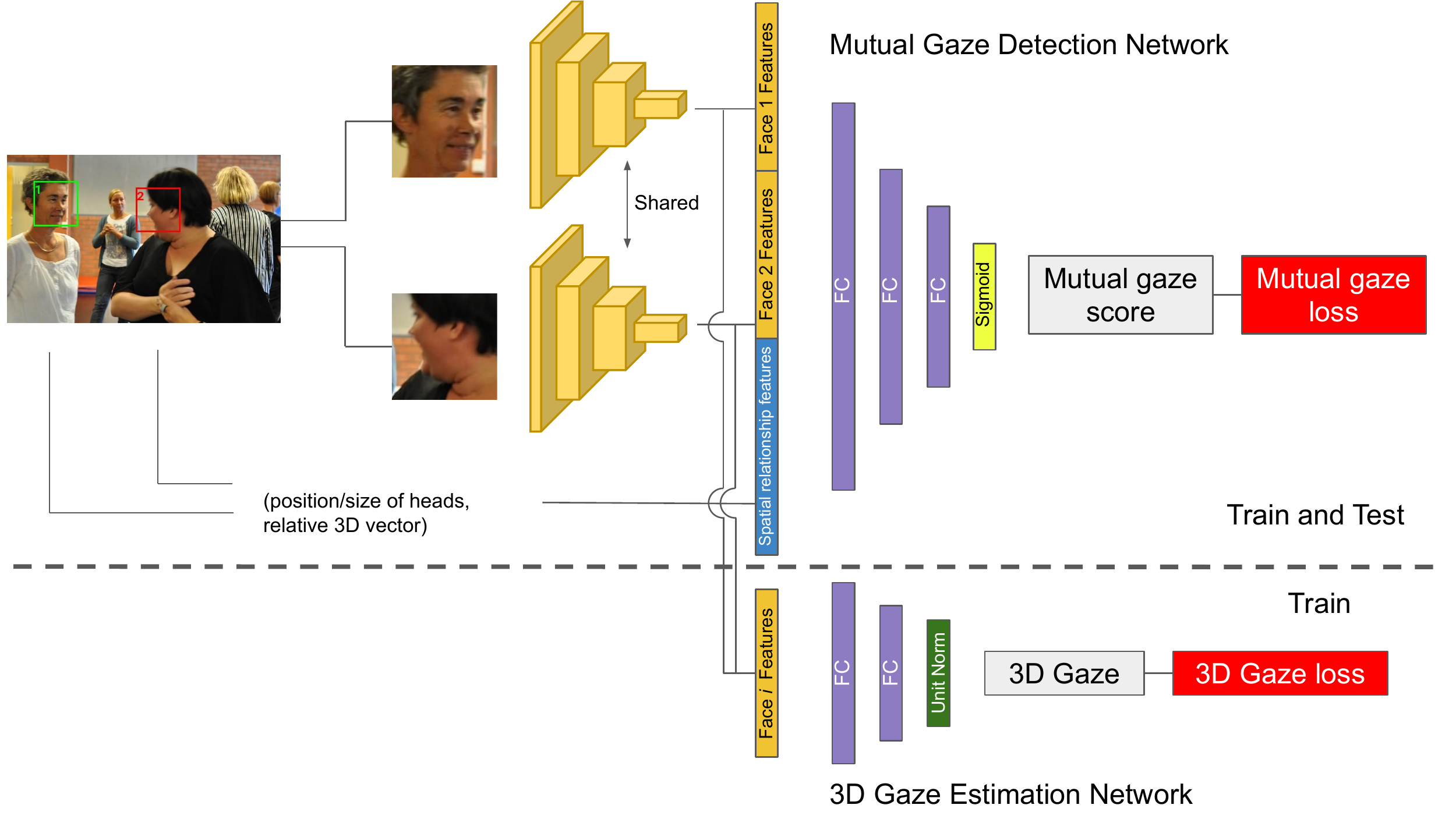}
\end{center}
   \caption{Proposed approach that consists of a primary mutual gaze detection branch which is used during both training and inference, and an auxiliary 3D gaze estimation branch which is used only during training.}
\label{fig:network_architecture}
\vspace{-3mm}
\end{figure*}

The proposed approach (Fig.~\ref{fig:network_architecture}) consists of four components: (i) Head image encoding, (ii) Head spatial encoding, (iii) mutual gaze detection, and (iv) 3D gaze estimation. The mutual gaze detection component takes the image and spatial encodings of the two head bounding boxes as input and produces a binary label as output. The 3D gaze estimation component takes a head image encoding as input and estimates the 3D gaze direction of the corresponding head. While mutual gaze detection is the primary task of interest, we also use an auxiliary 3D gaze estimation task during training to learn a better head image encoder. Thanks to the power of multi-task learning, this leads to improved mutual gaze detection performance. The entire network is trained using a dataset in which each data sample consists of a pair of head bounding boxes in an image along with a binary mutual gaze label. For training the 3D gaze estimation component, we use pseudo 3D gaze labels generated from positive mutual gaze samples.
\subsection{Head image encoding}
We use a shared CNN to encode each head image ($64 \times 64$) into a 12-dimensional feature representation. Our encoder consists of four convolutional layers that follow the architecture specification (kernel size, number of channels and stride) presented in~\cite{Marin19a}. However, since this work focuses on image-based detection, we use 2D convolutions instead of 3D convolutions. These head image features are used as input by both the mutual gaze and 3D gaze components.

\subsection{Head spatial encoding}
\label{sec:spatial_encoding}
 Apart from head image encodings, we also use features that encode the spatial information (in both 2D and 3D) of the two head bounding boxes.
 \subsubsection{2D spatial encoding:} Our 2D spatial encoding is an 8-dimensional feature vector consisting of normalized center coordinates (x and y coordinates) and sizes (width and height) of the two bounding boxes. We normalize them by dividing by the maximum of image height and width. 

 \subsubsection{3D spatial encoding:}
\begin{figure}[t]
\begin{center}
   \includegraphics[trim={0cm 1cm 0cm 0cm},width=0.48\textwidth]{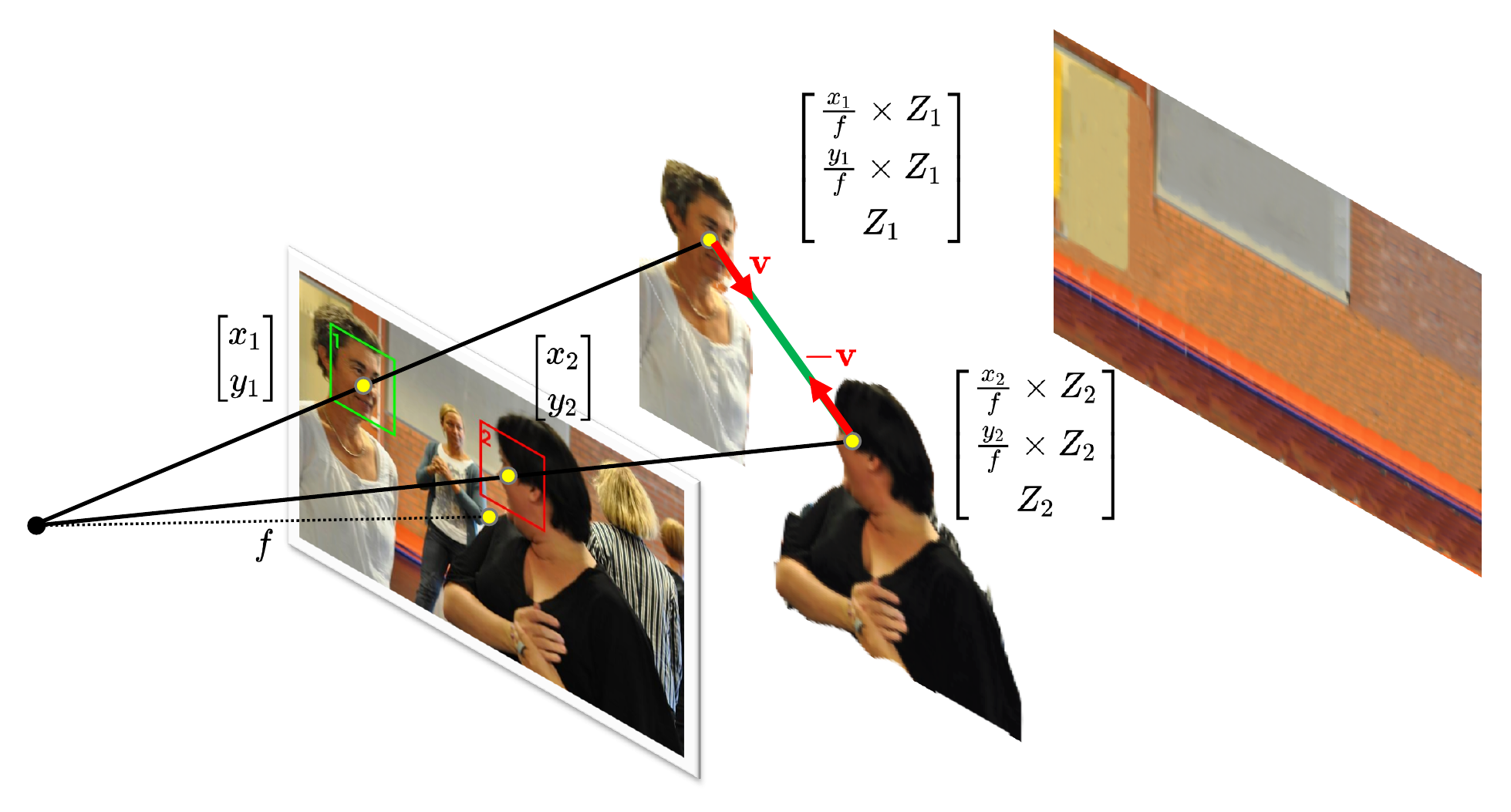}
\end{center}
   \caption{The location of head centers in the image and the corresponding 3D locations based on focal length $f$, and depth estimates $Z_1$ and $Z_2$. The vectors $\mathbf{v}$ and $-\mathbf{v}$ represent the pseudo 3D gaze labels for the first and second subjects, respectively.}
\label{fig:camera_laeo}
\vspace{-3mm}
\end{figure}

When two people are looking at each other, their 3D gaze directions roughly align with the direction of the relative 3D position vector between their head centers (see Fig.~\ref{fig:camera_laeo}). Since the 3D gaze information is already available to the mutual gaze detection component (indirectly) through head image encodings, providing the direction of the relative 3D head position vector as an additional input feature can be beneficial for mutual gaze detection.

Under the pinhole camera projection model (Fig.~\ref{fig:camera_laeo}), the camera-centered 3D coordinates of the $i^{th}$ head center can be estimated as
\begin{align}
\left[\frac{x_i}{f} Z_i, \frac{y_i}{f} Z_i, Z_i \right],
\end{align}
where $f$ is the camera focal length, $(x_i, y_i)$ are the image coordinates of the head bounding box center with origin at the image center, and $Z_i$ is the distance of the head center from the camera center along the optical axis. We assume that the principal point of the camera lies in the center of the image. 

For a typical image downloaded from the internet, we rarely have access to the focal length and depth information needed to estimate the 3D head center coordinates. Hence, we use approximate estimates for both of them. The focal length is approximated using the maximum of image width and height. This estimate roughly corresponds to $53^{\circ}$ horizontal Field of View (FoV):
\begin{equation}
    f = \frac{\text{max}(w, h)}{2} \text{cot}\left(\frac{53^{\circ}}{2} \right) \approx \text{max}(w, h).
\end{equation}

Motivated by~\cite{moon2019camera}, we assume that the depth of a head is inversely proportional to the square root of the corresponding 2D bounding box area: 
\begin{align}
Z_i = \frac{\alpha}{\sqrt{A_i}},
\end{align}
where $\alpha$ is a proportionality constant and $A_i$ denotes the 2D bounding box area of the $i^{th}$ head. This is a coarse approximation since it assumes that the heads of different individuals are of the same size and the head detector gives the same size bounding boxes for different individuals at the same depth. Using the above approximations, we compute the direction $\mathbf{v}$ of the relative 3D head position vector as
\begin{equation}
\begin{aligned}
& {\mathbf{v}}  =\frac{[\check{\mathbf{x}}, \check{\mathbf{y}}, \check{\mathbf{z}}] \,\, }{ \Vert [\check{\mathbf{x}}, \check{\mathbf{y}}, \check{\mathbf{z}}] \Vert_2},\  \check{\mathbf{x}} = \frac{1}{f}\left(\frac{x_2}{\sqrt{A_2}} - \frac{x_1}{\sqrt{A_1}}\right),\\
& \check{\mathbf{y}} = \frac{1}{f}\left(\frac{y_2}{\sqrt{A_2}} - \frac{y_1}{\sqrt{A_1}}\right),\ \check{\mathbf{z}} = \frac{1}{\sqrt{A_2}} - \frac{1}{\sqrt{A_1}} .
\label{eqn:relative_position}
\end{aligned}
\end{equation}
Note that the proportionality constant $\alpha$ does not appear here because it only contributes to the magnitude of the relative 3D head position vector and not to the direction.

The overall spatial encoding is an 11-dimensional feature vector that consists of 8-dimensional 2D encoding and 3-dimensional 3D encoding. Though our 3D spatial encoding is based on approximate depths and focal lengths, our experimental results show that it improves the performance of mutual gaze detection significantly (see Table~\ref{tab:ablation_results}).
\subsection{Mutual gaze detection}
For mutual gaze detection, we use a network that consists of three fully-connected layers with 16, 8 and 1 output nodes, respectively. The input to the network is a 35-dimensional feature vector obtained by concatenating the left head image encoding, right head image encoding and head spatial encodings. The first two layers use ReLU non-linearity, and the last layer uses sigmoid function to generate mutual gaze score in $[0, 1]$ range. 

\subsection{3D gaze estimation}
For 3D gaze estimation, we use a network that consists of two fully-connected layers with 6 and 3 output nodes, respectively, followed by a unit normalization layer. The input to this network is the 12-dimensional image encoding of a head. The first layer uses ReLU non-linearity and the second layer is a linear layer.
 
\subsection{Training and inference}
\label{sec:training}
While we have ground truth mutual gaze labels in our training dataset, we do not have ground truth 3D gaze labels. Earlier, we described how we estimate the direction $\mathbf{v}$ of the relative 3D head position vector. When two people are looking at each other, i.e, a positive mutual gaze sample, their 3D gaze directions roughly align with the relative 3D head position vector $\mathbf{v}$. Hence, for positive mutual gaze samples, we use $\mathbf{v}$ and $-\mathbf{v}$ as pseudo 3D gaze labels for the first and second heads, respectively. 

We train the entire network (head image encoder, mutual gaze detection network and 3D gaze estimation network) in an end-to-end fashion starting from randomly initialized weights using two loss functions: binary cross entropy loss $\mathcal{L}_{bce}$ on the mutual gaze detection scores and $\mathcal{L}_2$ loss on the 3D gaze direction estimates. While the mutual gaze detection loss is used for both positive and negative mutual gaze samples, the 3D gaze estimation loss is used only for positive mutual gaze samples (using the pseudo 3D gaze labels). The overall loss function used for training is given by
\begin{align}
\mathcal{L}  = \mathcal{L}_{bce}(\hat{l}, l) +  \lambda    l  \left(  \frac{\mathcal{L}_2(\hat{\mathbf{g}}_1, \mathbf{v}) + \mathcal{L}_2 (\hat{\mathbf{g}}_2, -\mathbf{v})}{2} \right),
\label{eqn:loss}
\end{align} % \mathds{1}
where $l \in \{0, 1\}$ is the binary mutual gaze label, $\hat{l} \in [0, 1]$ is the predicted mutual gaze score, $\hat{\mathbf{g}}_1$ and $\hat{\mathbf{g}}_2$ are the 3D gaze direction estimates for the first and second heads, respectively, and $\lambda$ is a hyper-parameter to balance the importance of two loss functions.

During inference, we ignore the 3D gaze network and use the output of the mutual gaze detection network.

\section{Experimental Results}
\begin{table*}[t]
\centering
\footnotesize{
\caption{Performance comparison with single-frame LAEO-Net.}
\label{tab:laeo_dataset_results}
\begin{tabular}{l l l l}
\hline
Training dataset & OI-MG & AVA-LAEO & UCO-LAEO\\
Test dataset &  OI-MG &  AVA-LAEO & UCO-LAEO\\\hline
Single-frame LAEO-Net  &   59.8 & 70.2 & 55.9\\
Proposed approach   &  \textbf{70.1} ($\uparrow$ 10.3)  & \textbf{72.2} ($\uparrow$ 2.0)  & \textbf{65.1} ($\uparrow$ 9.2)\\\hline
\end{tabular}\\\vspace{-3mm}
}
\end{table*}
\subsection{Datasets and evaluation metric}
We use three image-based mutual gaze detection datasets in our experiments, namely UCO-LAEO, AVA-LAEO and Open-Images-MG (OI-MG).
\subsubsection{UCO-LAEO~\cite{Marin19a}:} This dataset consists of 36,358 labeled pairs of humans extracted from 129 short video clips derived from TV shows. It is divided into training and test sets consisting of 32,500 and 3,858 pairs, respectively. This dataset comes with head bounding boxes and we directly use these boxes in our experiments.

\subsubsection{AVA-LAEO~\cite{Marin19a}:} This dataset consists of 172,330 labeled pairs of humans extracted from 50,797 video frames of the AVA dataset~\cite{AVADataset}. It is divided into training and test sets consisting of 137,976 and 34,354 pairs, respectively. This dataset comes with a head detector and we use this detector for generating head bounding boxes for this dataset.
\begin{figure}[t]
\begin{center}
\includegraphics[trim= 0 150 30 0,clip,width = 0.48\textwidth]{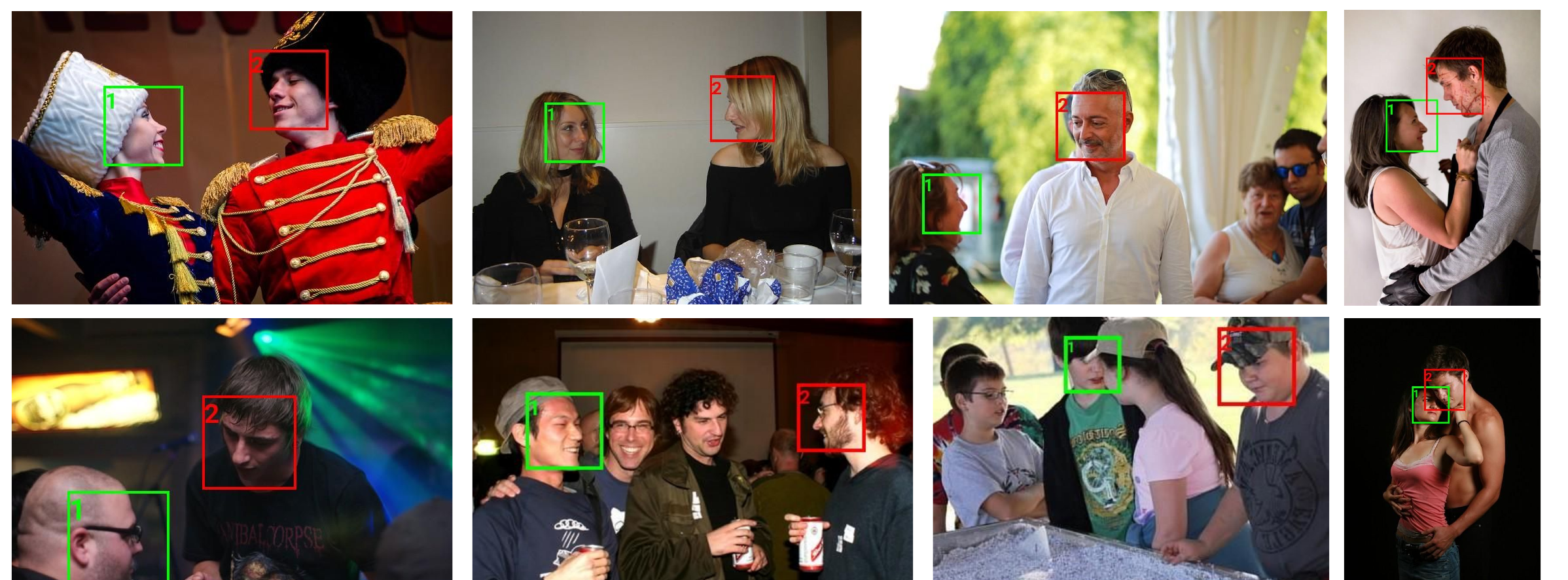}
\end{center}
   \caption{OI-MG dataset: Top and bottom rows show some positive and negative mutual gaze samples, respectively.}
\label{fig:oi_laeo}
\vspace{-3mm}
\end{figure}

\subsubsection{OI-MG:}This is a new dataset collected by us. UCO/AVA-LAEO datasets contain video frames collected from TV series and movies which are captured using cinematic lenses. In contrast, OI-MG dataset consists of in-the-wild still images taken by everyday photographers who use photography equipment (e.g. mobile devices) that is quite different from cinematography equipment. It has 33,069 labeled pairs of heads (8,972 positive and 24,097 negative pairs) extracted from 29,186 images of Open Images dataset~\cite{Alina2018OpenImages}. We used our implementation of SSD~\cite{LiuAESRFB16} detector for detecting human heads. Each pair of heads was annotated by five different annotators, and the ground truth label was chosen based on majority voting. Figure~\ref{fig:oi_laeo} shows some sample images from this dataset. This dataset is divided into 26,410 training pairs and 6,659 test pairs with zero overlap between training and test images.
\subsubsection{Evaluation metric:} Since this is a binary classification problem, we use the area under precision-recall curve, i.e., Average Precision (AP) as the metric.
\subsection{Training}
In all our experiments, the value of loss function parameter $\lambda$ (equation~\eqref{eqn:loss}) was set to 1.0. All the models were trained using RMSprop optimizer with mini-batches of 128 samples. We used an initial learning rate of $5 \times 10^{-4}$ which was gradually reduced by a factor of 0.94 after every 100K steps until convergence. To increase the robustness of the models, during training, we randomly applied horizontal flipping to the images, jittered the head bounding boxes, and adjusted the image intensity and contrast.

\begin{figure}[t]
\begin{center}
\includegraphics[trim= 0 10 130 0,clip,width=8.2cm, height=5.1cm]{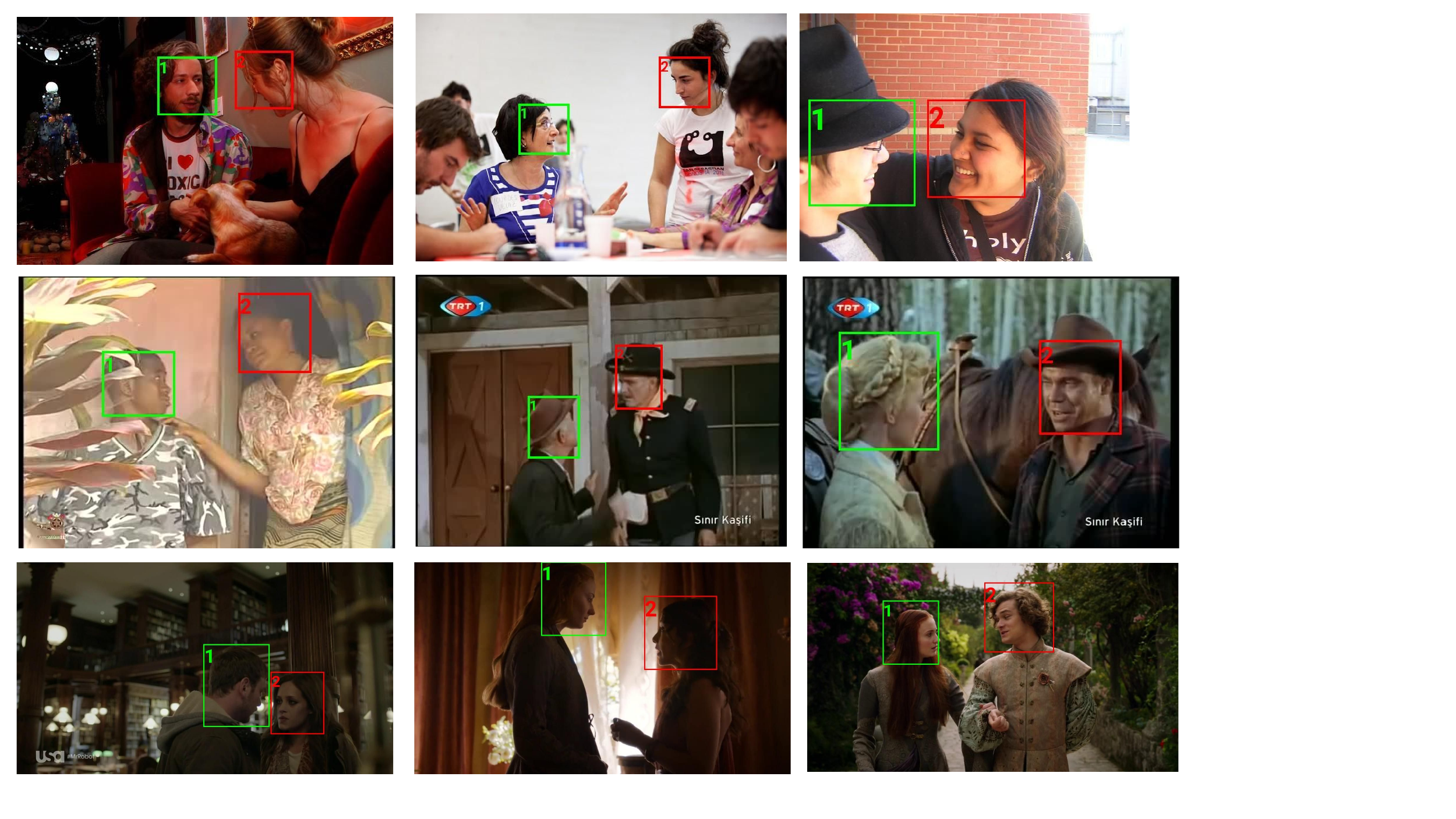}
\end{center}
\vspace{-3mm}
   \caption{Positive mutual gaze samples from OI-MG (first row), AVA-LAEO (second row), and UCO-LAEO (third row) datasets for which the proposed approach predicted high scores (above 0.75). The proposed approach can detect mutual gaze under various different combinations of head poses and positions.}
\label{fig:laeo_qualitative_results_high}
\vspace{-3mm}
\end{figure}

\begin{figure}[t]
\begin{center}
\includegraphics[trim= 0 10 130 0,clip,width=9.2cm, height=5.3cm]{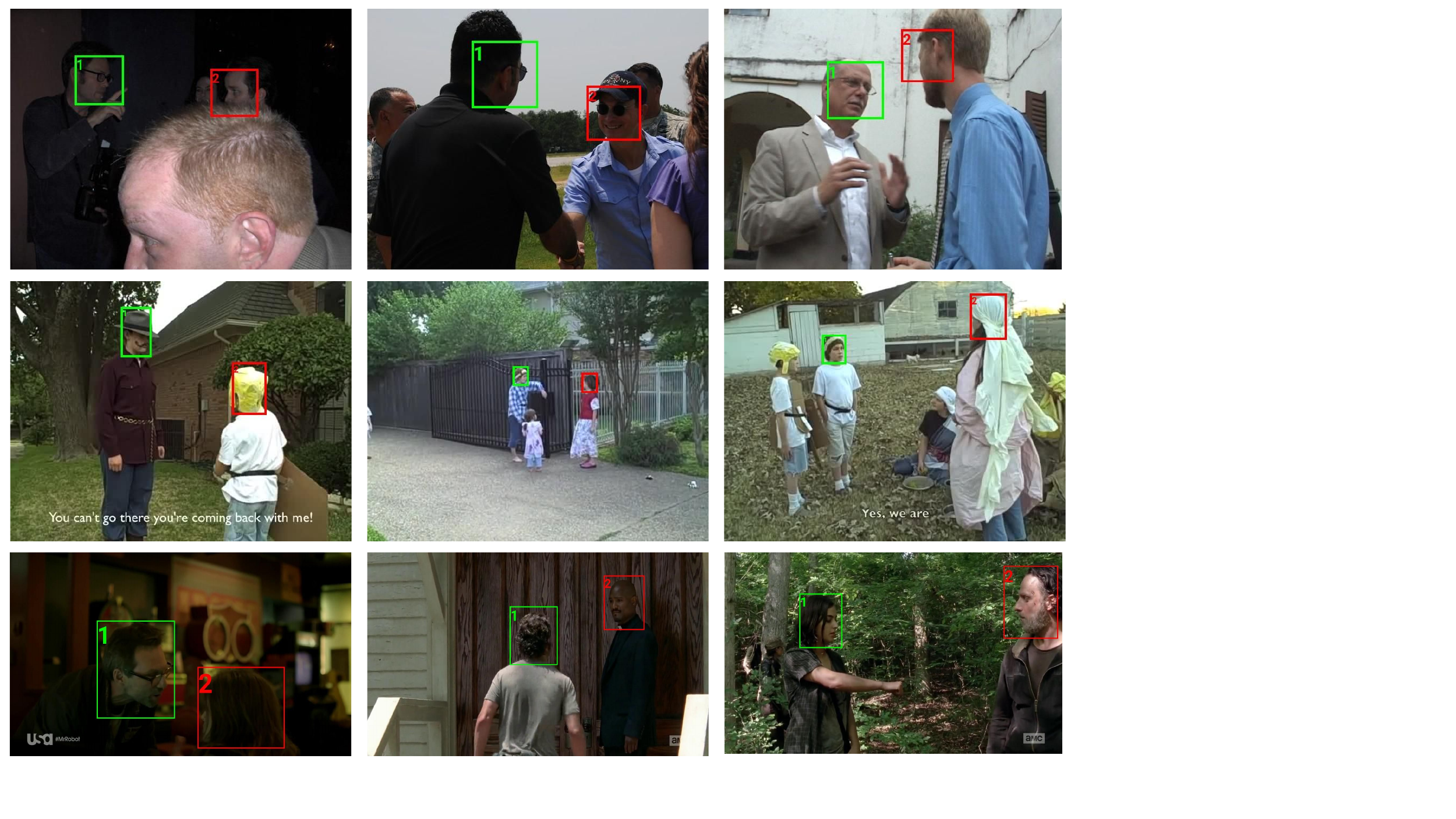}
\end{center}
\vspace{-5mm}
   \caption{Positive mutual gaze samples from OI-MG (first row), AVA-LAEO (second row), and UCO-LAEO (third row) datasets for which the proposed approach predicted low scores (below 0.25). Common failure reasons are occlusions, low contrast, bounding box drifts, low resolution and headwear.}
\label{fig:laeo_qualitative_results_low}
\vspace{-1mm}
\end{figure}

\subsection{Current state-of-the-art approach}
The current state-of-the-art approach for mutual gaze detection is the LAEO-Net model proposed in~\cite{Marin19a}. The original LAEO-Net model is designed for video-based mutual gaze detection, and uses 3D convolutions for spatio-temporal modeling. Since this work focuses on image-based mutual gaze detection, we converted the video LAEO-Net into a single-frame LAEO-Net by replacing the 3D convolutions with 2D convolutions. For fair comparison, similar to the proposed approach, we trained single-frame LAEO-Net in an end-to-end fashion starting from randomly initialized weights. 

\begin{table*}[t]
\centering
\footnotesize{
\caption{Contributions of 3D gaze estimation task and 3D spatial encoding.}
\label{tab:ablation_results}
\begin{tabular}{l l l l }
\hline
  Training dataset & OI-MG  & AVA-LAEO & UCO-LAEO  \\\hline
  Test dataset & OI-MG  & AVA-LAEO & UCO-LAEO  \\\hline
  Proposed approach  & \textbf{70.1}  & \textbf{72.2}  & \textbf{65.1}  \\
  Proposed approach without 3D gaze task & 67.1 ($\downarrow$ 3) & 68.3 ($\downarrow$ 3.9)  & 62.9 ($\downarrow$ 2.2)  \\ 
  Proposed approach without 3D spatial encoding & 68.0 ($\downarrow$ 2.1) & 70.1 ($\downarrow$ 2.1)  & 61.6 ($\downarrow$ 3.5)  \\
  Proposed approach without 3D gaze and 3D spatial encoding & 66.4 ($\downarrow$ 3.7) & 66.9 ($\downarrow$ 5.3)  & 60.4   ($\downarrow$ 4.7) \\
  \hline
\end{tabular}
\vspace{-2mm}
}
\end{table*}
\subsection{Comparison with image-based state of the art}

Table~\ref{tab:laeo_dataset_results} compares the proposed approach with single-frame LAEO-Net on three image datasets. On all three datasets, the proposed approach outperforms single-frame LAEO-Net
%significantly
increasing the AP by 10.3 points for OI-MG, 9.2 points for UCO-LAEO and 2 points for AVA-LAEO. Compared to single-frame LAEO-Net, the proposed approach is able to utilize the auxiliary 3D gaze estimation task to learn a better head image encoder, and also  make use of the 3D information provided by the relative 3D head position vector resulting in improved mutual gaze detection performance.

\begin{figure}[t]
\begin{center}
\includegraphics[trim= 0 10 70 0,clip,width = 0.5\textwidth]{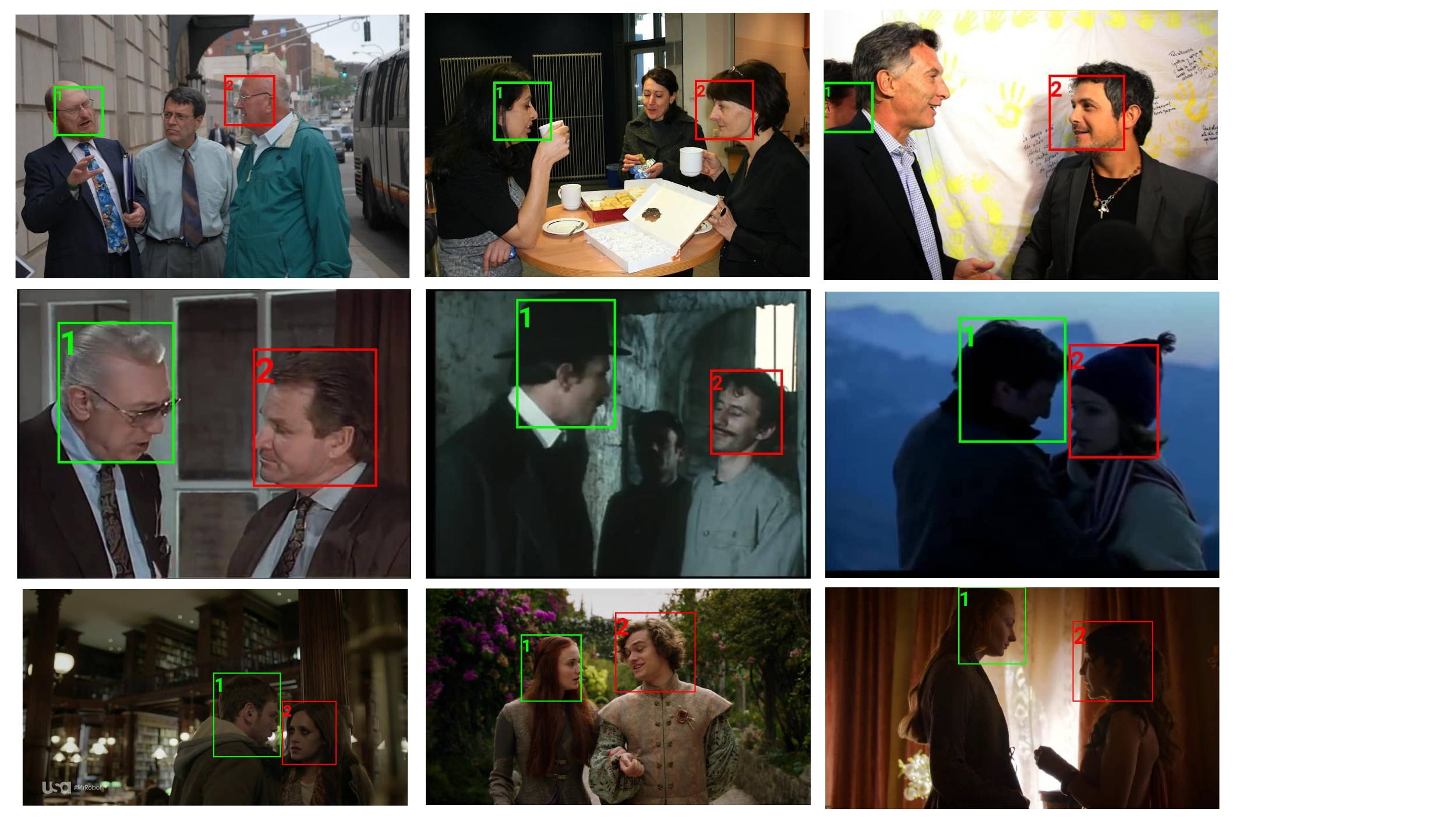}
\end{center}
\vspace{-2mm}
   \caption{Negative mutual gaze samples from OI-MG (first row), AVA-LAEO (second row), and UCO-LAEO (third row) datasets for which the proposed approach predicted high scores (above 0.75). Common failure reasons are bad lighting conditions, occlusions and closed eyes (e.g., the second and the third samples in the third row).}
\label{fig:laeo_qualitative_results_neg}
\vspace{-3mm}
\end{figure}

Figure~\ref{fig:laeo_qualitative_results_high} shows some positive mutual gaze samples for which the proposed approach predicts high scores (above 0.75). We can see that the proposed approach is able to handle a wide variety of head pose and position combinations. Figure~\ref{fig:laeo_qualitative_results_low} shows some positive mutual gaze samples for which the proposed approach predicts low scores (below 0.25). We can see that most of the failure cases are due to occlusions, bad lightning conditions, headwear, low resolution and bounding box drifts. Figure~\ref{fig:laeo_qualitative_results_neg} shows some negative mutual gaze samples for which the proposed approach predicts high scores (above 0.75). We can see that most of the failure cases are due to closed eyes, occlusions, and bad lightning conditions. Our head encoder does not specifically focus on eyes and hence may not be capturing enough details to detect closed eyes in some cases. Using eye image patches as inputs to the encoder along with head image patches could potentially alleviate this problem.

\subsection{Contributions of auxiliary 3D gaze estimation task and 3D spatial encoding} \label{sec:ablation}
Table~\ref{tab:ablation_results} shows the results of the proposed approach with and without the auxiliary 3D gaze estimation task and 3D spatial encoding. Removing the auxiliary 3D gaze estimation task reduces the AP by 3, 3.9 and 2.2 points on OI-MG, AVA-LAEO and UCO-LAEO, respectively. Removing the 3D spatial encoding reduces the AP by 2.1 points on both OI-MG and AVA-LAEO, and 3.5 points on UCO-LAEO. Removing both reduces the AP by 3.7, 5.3 and 4.7 points on OI-MG, AVA-LAEO and UCO-LAEO, respectively. These results clearly demonstrate that both the auxiliary 3D gaze estimation task and 3D spatial encoding contribute
%significantly
to the final performance and they are complementary to each other. To further demonstrate the effectiveness of the 3D gaze estimation task and 3D spatial encoding, we added them to single-frame LAEO-Net. This model augmentation increased the AP by 2.7, 1.4 and 0.7 points on OI-MG, AVA-LAEO and UCO-LAEO, respectively.

\subsection{Comparison under external training data setting}
Since UCO-LAEO is a small dataset with limited diversity (129 shots from 4 TV shows),~\cite{Marin19a} used additional synthetic data generated from AFLW dataset ~\cite{KostingerWRB11} for training to improve the results on UCO-LAEO. As we do not have access to the synthetic data used by~\cite{Marin19a}, we evaluate AVA-LAEO (which is much larger than UCO-LAEO) trained models on UCO-LAEO  for comparison under external training data setting. Table~\ref{tab:external_data_results} shows the corresponding results. In this setting, the proposed approach outperforms single-frame LAEO-Net by 8.0 points. The last column presents the only single-frame LAEO-Net result reported in~\cite{Marin19a}. The proposed approach (when trained on AVA-LAEO) outperforms this result by 3.5 points.\footnote{We acknowledge that this is not a fair comparison. However, we do not have access to the AFLW synthetic data used by~\cite{Marin19a}.}

We also evaluated the proposed model trained on AVA-LAEO on our OI-MG dataset. While this AVA-LAEO model generalized well to UCO-LAEO, it failed to generalize to OI-MG. It achieved an AP of 32.8, which is significantly lower than 70.1 achieved by the model trained on OI-MG itself. We conjecture that the reason for this is the difference in the data distributions of OI-MG and AVA/UCO-LAEO datasets. OI-MG consists of still images downloaded from Flickr and is significantly different from AVA/UCO-LAEO which are video frame datasets created from movie clips. This emphasizes the need for a new dataset for LAEO problem.
\begin{table}
\setlength{\tabcolsep}{5pt}
\centering
\footnotesize{
\caption{Comparison under external training data setting.}
\label{tab:external_data_results}
\begin{tabular}{l l l}
\hline
Training dataset & AVA-LAEO & UCO-LAEO + AFLW\\
Test dataset & UCO-LAEO & UCO-LAEO\\\hline
Single-frame LAEO-Net & 68.2 & 72.7$^{*}$\\
Proposed approach   & \textbf{76.2} ($\uparrow$ 8.0) & -\\\hline
\end{tabular}\\\vspace{1mm}
*Number reported in~\cite{Marin19a}
}
\end{table}
\begin{table*}[t]
\renewcommand{\arraystretch}{1.5}
\centering
\footnotesize{
\caption{Performance for different amounts of training data.}
\label{tab:varying_amount_of_data}
\begin{tabular}{l l l l l}
\hline
Training data percentage & 100\% & 50\% & 25\% & 12.5\%\\\hline
\multicolumn{5}{c}{Training dataset: AVA-LAEO, Test dataset: AVA-LAEO}\\\hline
Single-frame LAEO-Net & 70.2 & 65.6 & 55.7 & 50.7\\
Proposed approach & \textbf{72.2} ($\uparrow$ 2.0) & \textbf{68.4} ($\uparrow$ 2.8) & \textbf{63.0} ($\uparrow$ 7.3) & \textbf{58.2} ($\uparrow$ 7.5) \\\hline
\multicolumn{5}{c}{Training dataset: OI-MG, Test dataset: OI-MG}\\\hline
Single-frame LAEO-Net & 59.8 & 53.5 & 44.9 & 42.5 \\
Proposed approach & \textbf{70.1} ($\uparrow$ 10.3) & \textbf{62.4} ($\uparrow$ 8.9) & \textbf{56.8} ($\uparrow$ 11.9) & \textbf{51.1} ($\uparrow$ 8.6)\\\hline

\end{tabular}\\
}
\end{table*}
\subsection{Comparison with video-based state of the art}
Different from the original LAEO-Net~\cite{Marin19a} which focuses on video-based detection, this work focuses on image-based mutual gaze detection.
%as images are captured more frequently than videos in real-world social interactions.
Since image-based detection and video-based detection are completely different tasks, it is not fair to compare image-based methods with video-based methods. That being said, the proposed image-based approach significantly outperforms (AP 72.2 vs. 50.6) the best video-based result from~\cite{Marin19a} on AVA-LAEO. On UCO-LAEO, our image-based result is comparable (AP 76.2 vs. 79.5) to the best video-based result from~\cite{Marin19a}.

\subsection{Performance in limited data setting}
To see how the performance gap between the proposed approach and single-frame LAEO-Net varies with the amount of training data, we trained both models by varying the number of training samples. The corresponding results are shown in Table~\ref{tab:varying_amount_of_data}. The proposed approach consistently outperforms single-frame LAEO-Net for different amounts of training data. Specifically, on AVA-LAEO, which is a large dataset with around 138K training samples, the performance gap is 
%significantly
higher for 12.5\% and 25\% training data when compared to 50\% and 100\% training data. This is because, when the amount of training data is limited, the proposed auxiliary 3D gaze estimation task provides additional training signal which helps the model to train well. Since OI-MG is a relatively smaller dataset with only 26.5K training samples, the performance gap is significant even when using 100\% data.

\subsection{Sensitivity to field of view}
\begin{figure}[t]
\begin{center}
\includegraphics[clip,width = 0.49\textwidth]{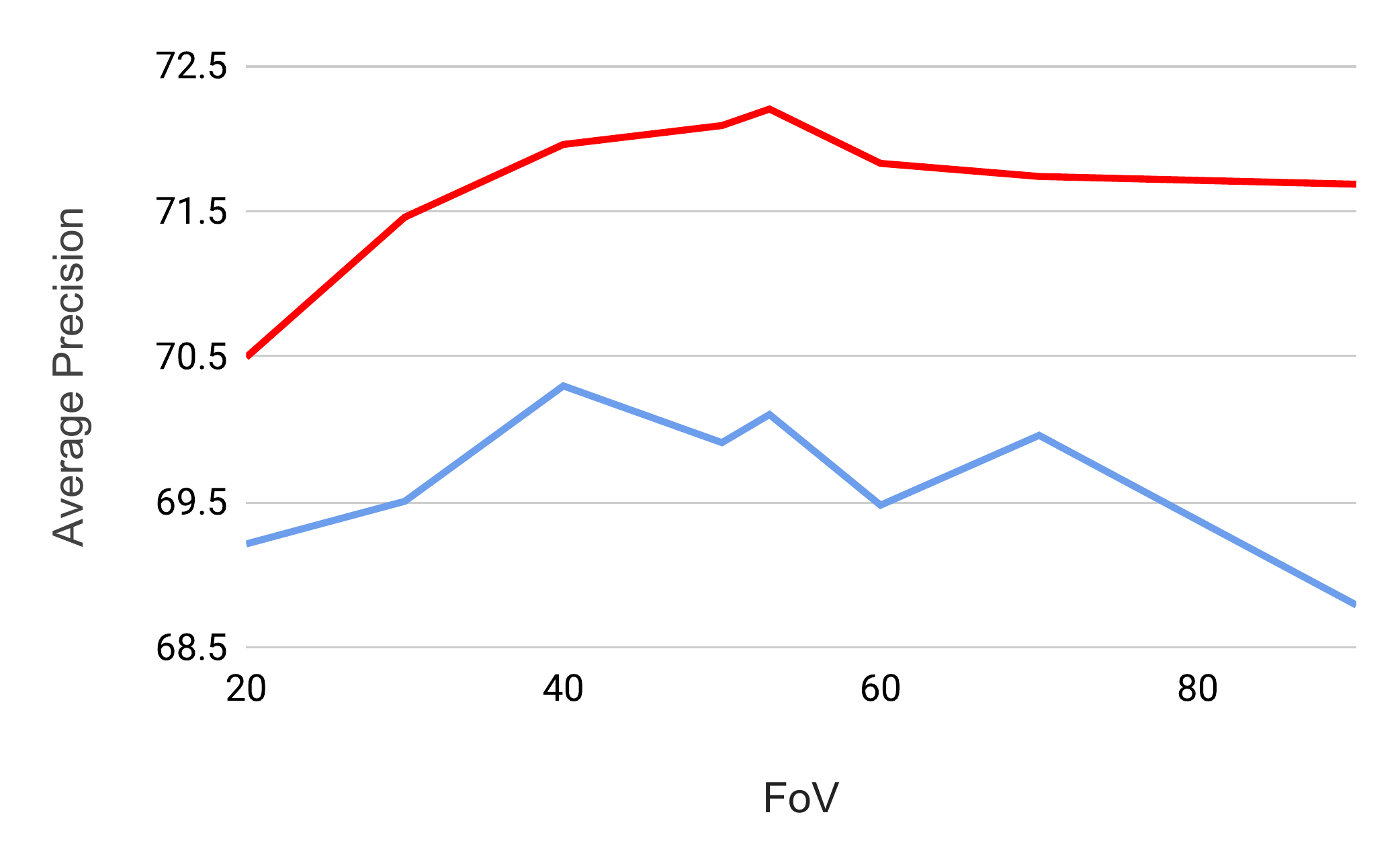}
\end{center}
\vspace{-5mm}
   \caption{Performance for different values of horizontal field of view. The red and blue curves correspond to AVA-LAEO and OI-MG, respectively.}
\label{fig:fov_results}
\vspace{-3mm}
\end{figure}

Since we do not know the focal lengths for images used in our experiments, we used the maximum of image width and height as the focal length. This estimate corresponds to $53^{\circ}$ horizontal FoV. Figure~\ref{fig:fov_results} shows how the performance varies with the value of horizontal FoV used to compute the focal length. The performance increases as we increase the FoV up to a certain value and then it starts to decrease. However, the results are fairly stable within $15^{\circ}-20^{\circ}$ range around the peak values. Different datasets have different best FoV values as the cameras used to capture can vary across datasets.

\section{Conclusions and Future Work}

In this work, we focused on detecting whether or not two people in an image are looking at each other. We proposed a network architecture that consists of a primary mutual gaze detection branch and an auxiliary 3D gaze estimation branch. The auxiliary 3D gaze estimation branch is trained using pseudo 3D gaze labels deduced from mutual gaze labels without the need for 3D gaze ground truth. Since both branches share the head image encoder, it allows the network to learn a robust head image encoding which leads to improved mutual gaze detection performance. We also proposed to use geometry-inspired 3D relative head position vector as additional input feature, and showed that it substantially improves the performance of mutual gaze detection. We empirically demonstrated that our approach outperforms the state-of-the-art single-frame LAEO-Net approach on three image datasets. Specifically, the proposed approach fares better when the amount of training data is limited.

Some of our failure cases (Figure~\ref{fig:laeo_qualitative_results_neg}) suggest that we need to pay extra attention to the eye regions while encoding the head image patches. We plan to explore models that explicitly process eye patches in the future.
%In this work, 
We generated pseudo 3D gaze labels
%for positive mutual gaze samples
and used 3D gaze estimation as an auxiliary task during training.
A positive mutual gaze label directly provides ground truth annotations for the VFA task since the gaze location is known. In the future, we plan to experiment with using an auxiliary VFA task during training. 

\section{Ethical and Broader Impact}
This work focuses on detecting if two people in an image are looking at each other or not. This research influences applications in social interaction detection and scene understanding. This work also introduces a new mutual gaze dataset. This dataset is based on the publicly-available Open Images dataset and does not include any private data. We do not foresee any negative ethical or societal effects for this work. We believe that making our dataset available to other researchers will encourage an open research environment.

%\clearpage

%\bibliographystyle{aaai21}
\bibliography{laeo_bib}

\end{document}